\documentclass{article} 
\usepackage{iclr2020_conference,times}

\usepackage{amsmath,amsfonts,bm}

\def\eqref#1{equation~\ref{#1}}

\def\1{\bm{1}}

\DeclareMathAlphabet{\mathsfit}{\encodingdefault}{\sfdefault}{m}{sl}
\SetMathAlphabet{\mathsfit}{bold}{\encodingdefault}{\sfdefault}{bx}{n}

\usepackage[utf8]{inputenc} 
\usepackage[T1]{fontenc}    
\usepackage{hyperref}       
\usepackage{url}            
\usepackage{booktabs}       
\usepackage{amsfonts}       
\usepackage{nicefrac}       
\usepackage{microtype}      
\usepackage{comment}
\usepackage{graphicx}

\title{What can human minimal videos tell us about dynamic recognition models?}

\author{Guy Ben-Yosef$^{1,2,5}$, Gabriel Kreiman$^{2,3}$, Shimon Ullman$^{2,4}$ \\
	$^1$Computer Science and Artificial Intelligence Laboratory , Massachusetts Institute of Technology, USA \\	
	$^2$Center for Brains, Minds, and Machines, Massachusetts Institute of Technology, USA \\	
	$^3$Children's Hospital, Harvard Medical School, USA \\
	$^4$Department of Computer Science and Applied Mathematics,	Weizmann Institute of Science, Israel\\
	$^5$GE Research, Artificial Intelligence, USA \\	
	\texttt{gby@csail.mit.edu, gabriel.kreiman@tch.harvard.edu, shimon.ullman@weizmann.ac.il}
}

\iclrfinalcopy 
\begin{document}

\maketitle

\begin{abstract}
  In human vision objects and their parts can be visually recognized from purely spatial or purely temporal information but the mechanisms integrating space and time are poorly understood. Here we show that human visual recognition of objects and actions can be achieved by efficiently combining spatial and motion cues in configurations where each source on its own is insufficient for recognition. This analysis is obtained by identifying minimal videos: these are short and tiny video clips in which objects, parts, and actions can be reliably recognized, but any reduction in either space or time makes them unrecognizable. State-of-the-art deep networks for dynamic visual recognition cannot replicate human behavior in these configurations. This gap between humans and machines points to critical mechanisms in human dynamic vision that are lacking in current models. 

\end{abstract}

\section{The role of temporal information in dynamic visual recognition}

Previous behavioral work has shown that human visual recognition can be achieved on the basis of spatial information alone ~\cite{Potter_etal_1969_JEP,Ullman_etal_2016_PNAS}, and on the basis of motion information alone, as in the case of identifying human activities from biological motion~\cite{Johansson_1973_PP}. At the neurophysiological level, neurons have been identified that respond selectively to objects and events based on purely spatial information, or motion information alone ~\cite{Sary_etal_1993_Science,Perrett_etal_1985_BBR}. 
However, several behavioral studies have also provided strong support suggesting that a combination of spatial and temporal information can aid recognition, e.g.,~\cite{Parks_1965_AJP,Morgan_etal_1982_JEP}, but whether and how space and time may be integrated remain unclear.

One of the domains in which temporal information is particularly relevant is action recognition. While several approaches and computational models have been developed to recognize actions from videos -- and recent models are 
based on late or early integration of the spatial and temporal information processed by deep networks (e.g.,~\cite{Giese_Poggio_2003_NatureReviews,Tran_etal_2015_ICCV,Donahue_etal_2015_CVPR,Karpathy_etal_2014_CVPR,Simonyan_Zisserman_2014_NIPS,Hara_etal_2018_CVPR}) -- 
it remains unclear whether current models make an adequate and human-like use of spatiotemporal information. 
In order to evaluate the use of spatiotemporal integration by computational models, it is crucial to construct test stimuli that ‘stress test’ the combination of spatial and dynamic features. A difficulty with current efforts is that in many action recognition data sets (e.g.~\cite{Soomro_etal_2012,Kay_etal_2017}) high performance can be achieved by considering purely spatial information, 
and so they are not ideally set up to rigorously test spatiotemporal integration.

Here we developed a set of stimuli that can directly test the synergistic interactions of dynamic and spatial information, to identify spatiotemporal features that are critical for visual recognition and to evaluate current deep network architectures on these novel stimuli. 
We tested minimal spatiotemporal configurations (also referred to below as minimal videos), which are composed of a set of sequential frames (i.e., a video clip), in which humans can recognize an object and an action, but where further small reductions in either the spatial dimension (i.e., reduction by cropping or down sampling of one or more frames) or in the time dimension (i.e., removal of one or more frames from the video) turns the configuration unrecognizable, and therefore also uninterpretable, for humans. 

\section{A search for minimal videos}
\label{gen_inst}

\begin{figure}[t]
	\centering
	\includegraphics[width=12cm]{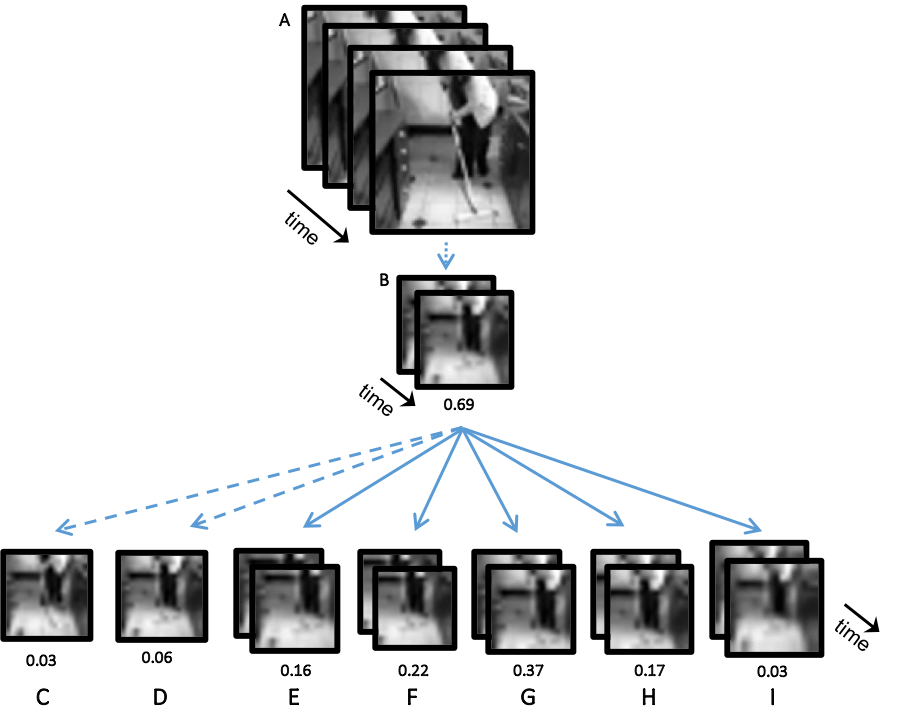}
    \vspace{-0.2cm} 
	\caption{
		{\bf Example of a minimal video.} 
		A short initial video clip showing ‘mopping’ activity {\bf(A)} was gradually reduced in both space and time to a minimal recognizable video configuration {\bf(B)}. The numbers on the bottom of each image show the fraction of subjects who correctly recognized the action (each subject saw only one of these images). The spatial and temporal trimming was repeated until none of the spatially reduced versions ({\bf E-I}, solid connections) or temporally reduced versions ({\bf C,D,} dashed connections) reached the recognition criterion of 50$\%$ correct answers. {\bf Spatial reduced versions:} In {\bf E} each frame was cropped in the top-right corner, leaving 80$\%$ of the original pixel size in {\bf B}. {\bf F,G,H} are similar versions where the crop is on the top-left, bottom-right, and bottom-left corners, respectively, {\bf I} is a version where the resolution of each frame was reduced to 80$\%$ of the frame in {\bf B}. {\bf Temporal reduced versions:} A single frame was removed, resulting in static frame\#1 in {\bf C}, and static frame\#2 in {\bf D}. 
		Follow 
		\href{https://www.dropbox.com/s/74d5kf8xqbfu24r/fig1.mp4?dl=0}{this link} 
		for the animated version of this figure. 
		}
	\label{fig:boat1}
\end{figure}

The search for each minimal video started from a short video clip, taken from the UCF101 dataset \cite{Soomro_etal_2012}, in which humans could recognize a human-object interaction. The search included different video snippets from various human-object interaction categories (e.g., ‘a person rowing’, ‘a person playing violin’, ‘a person mopping’, etc). 
The original video snippets were reduced to a manually selected 50x50 pixel square region, cropped from 2 to 5 sequential non-consecutive frames, and taken at the same positions on each frame. These regions served as the starting configurations in the search for minimal video configurations described below. In the default condition, frames were presented dynamically in a loop at a fixed frame rate of 2Hz. 
An illustration of the starting configuration for one of these examples is shown in Fig.~\ref{fig:boat1}A 
(We highly encourage the reader to view the dynamic version of Fig.~\ref{fig:boat1} that is attached to this submission).
The starting configuration was then gradually reduced in small steps of 20\% in size and resolution (same procedure as in a previous study \cite{Ullman_etal_2016_PNAS}). At each step, we created reduced versions of the current configuration, namely five spatially reduced versions decreasing in size and resolution, as well as temporally reduced versions where a single frame was removed from the video configurations. Each reduced version was then sent to Amazon’s Mechanical Turk (MTurk), where 30 human subjects were asked to freely describe the object and action. MTurk workers tested on a particular video configuration were not tested on additional configurations from the same action type. 
The success rates in recognizing the object and the action were recorded for each example. We defined a video configuration as recognizable if more than 50\% of the subjects described both the object and the action correctly. 

The search continued recursively for the recognizable reduced versions, until it reached a video configuration that was recognizable, but all of its reduced versions (in either space or time) were unrecognizable. We refer to such a configuration as a ‘minimal video’. An example of a minimal video is shown in Fig.~\ref{fig:boat1}B, and the reduced sub-minimal versions are shown in Fig.~\ref{fig:boat1}C-I. 
As shown in Fig.~\ref{fig:boat1}B-D, the spatial content of the minimal and temporal sub-minimal configurations was very similar: there were only minor spatial content added from frame 1 to frame 2. Despite the similarity of the two frames, there was a large difference in human recognition due to the removal of the motion signal. 
As shown in Fig.~\ref{fig:boat1}E-H, in the tested cases the motion content of the minimal and (spatial) sub-minimal is very similar, that is, the pixels that are cropped out do not remove a large amount of image motion. This implies that the motion signal alone is not a sufficient condition for human recognition of minimal videos. 

A prominent characteristic of minimal video configurations was a clear and consistent gap in recognition between the minimal configurations and their sub-minimal versions. The mean recognition rate was $0.71\pm 0.11 (mean\pm SD)$ for 20 minimal video configurations (such as the one in Fig. 1B), $0.29\pm0.15$ for the spatial sub-minimal configurations (such as the ones in Fig. 1E-I), and ($0.16 \pm 0.14$) for the temporal sub-minimal configurations (such as the ones in Fig. 1C-D). The difference in recognition rates between the minimal and sub-minimal configurations were statistically highly significant.
The minimal videos included 2 frames of n x n pixels, where $n=20 \pm 7.1$ on average. 
Although highly reduced in size, the recognition rate for the minimal videos was high, and was not too far from the recognition rate of the original UCF101 video clips (mean recognition = $0.94\pm 0.07$), even though the original clips had an average of 175 frames (versus 2 frames), each with 320x240 colored RGB pixels.
Recognition rates for the minimal videos were also close to the recognition rates for the level above it in the search tree (the ‘super minimal video configuration’, mean recognition = $0.81\pm0.07$). 

\section{Convolutional network models for recognizing minimal videos}
\label{sec:3D_2D_CNNs}

To further understand the mechanisms of spatiotemporal integration in recognition, we tested current models of spatiotemporal recognition on our set of minimal videos, and compared their recognition performance to human recognition. Our working hypothesis was that minimal videos require integrating spatial and dynamic features, which are not used by current models. The tested models included the C3D model~\cite{Tran_etal_2015_ICCV,Hara_etal_2018_CVPR}, the two-stream network model~\cite{Simonyan_Zisserman_2014_NIPS}, and the RNN-based model~\cite{Donahue_etal_2015_CVPR}, 
three popular network deep models that represent three different modern approaches to spatiotemporal recognition.
Our computational experiments included three types of tests with increasing amount of specific training, to compare human visual spatiotemporal recognition with existing models. 

{\bf Testing pre-trained DNNs on minimal videos:} In the first test, models were pre-trained on the UCF-101 dataset for video classification. We tested such pre-trained models on our set of minimal videos, to explore their capability to generalize from real-world video clips to minimal configurations. Classification accuracy by the C3D model for minimal videos was significantly lower than the classification accuracy achieved for the original full video clips, from which we cropped the minimal videos.
As example for a typical minimal video, 75\% of humans could correctly identified the action while the correct answer was not even among the top 10 for the network model.
%
%

{\bf Testing fine-tuned DNNs on minimal video:} Next, we evaluated whether training the models with minimal videos (fine-tuning) could help improve their performance. We used a binary classifier based on the convolutional 3D network model~\cite{Tran_etal_2015_ICCV,Hara_etal_2018_CVPR}, which was pre-trained on the large SportM dataset
~\cite{Karpathy_etal_2014_CVPR}. The network was then fine-tuned on a training set including 25 positive examples similar to a minimal video from a single category and type (the ‘rowing’ minimal video, see examples in Supplementary Fig.~\ref{fig:3D_2D_cnns}A, all positive examples were validated as recognizable to humans), as well as 10000 negative examples (e.g., Supplementary Fig.~\ref{fig:3D_2D_cnns}B). Data balancing techniques were used to ensure that the results would not be biased by the imbalance between positive and negative examples. The binary classifier was then tested on a novel set of 
positive and negative examples similar to the ones used during training. Since our set of positive examples was constrained to specific body parts and specific viewing positions in ‘rowing’ video clips, the fine-tuned classifier was able to correctly classify most of the negative test examples; the Average Precision (AP) was 0.94. Still, a non-negligible set of negative examples was given high positive score by the fine-tuned model, from which we composed a new set that we refer to as ‘hard negative video configurations’ for further analysis. The hard negative configurations included 30 negative examples that were erroneously labeled as positive by the fine-tuned network model. Comparing accuracy of human and network recognition for the set of hard negative configurations further revealed a significant gap: humans were not confused by any of the hard negative examples, while the fine-tuned network scored the hard negatives higher than most positive examples. 

{\bf Testing DNNs on minimal vs. sub-minimal videos:} A distinctive property of recognition at the minimal level is the sharp gap between minimal and sub-minimal videos. We therefore further compared recognition by the binary CNN classifier and human recognition by testing whether the network model was able to reproduce the gap in human recognition between the minimal configurations and their spatial and temporal sub-minimal ones. For this purpose, we collected a set of minimal and sub-minimal videos showing a large gap in human recognition, which did not overlap with the training set for the network model. We tested the fine-tuned network model from above on a set containing 20 minimal videos, 20 temporal sub-minimal configurations and 20 spatial sub-minimal configurations, all from the same category of ‘rowing’ in a similar viewing position and size.
Although this network model performed well on normal action classification (0.94 AP, see above) it was not able to replicate human recognition performance over this test set. While there was a clear gap in human recognition between minimal and spatial sub-minimal videos (average gap in human recognition rate 0.63), and between minimal and temporal sub-minimal videos (average gap in human recognition rate 0.68), the differences in recognition scores given by the network model for the minimal and sub-minimal examples were significantly smaller.
This discrepancy between human behavior and the models persisted even after using standard data augmentation techniques. 
\section{Spatiotemporal interpretation and future models}

Subjects who could recognize the minimal videos could also identify internal parts in them (e.g., body parts, action objects, etc.), and we hypothesized that action recognition is accompanied by detailed interpretation of the image parts as well as spatiotemporal relations between parts. To test this conjecture we ran a new series of experiments where subjects were instructed to describe internal components of the videos. MTurk subjects were presented with the minimal videos, along with a probe pointing to one of its internal spatial components. We defined a component as 'recognized' if it was correctly labeled by more than 50\% of the subjects. Average recognition for 31 components in 5 different types of minimal videos was 0.77$\pm$0.17. For example, in the “mopping” minimal video (Fig.~\ref{fig:boat1}) most of the subjects were able to correctly label the arm, legs, stick and vacuum.
To assess whether the dynamic video configurations were necessary for interpretation, we repeated the experiment using the spatial sub-minimal and temporal sub-minimal versions, using the same procedure of inserting a probe in the frames. In contrast to the reports obtained from the minimal videos, subjects consistently struggled to recognize the parts in the sub-minimal videos. 

We conclude that since minimal videos are limited in their amount of visual information, and require efficient use of the existing spatial and dynamic cues, comparing their recognition by humans and existing models uncovers differences in the use of the available information. 
Future studies could extend recent modeling of full interpretation of spatial minimal images~\cite{Ben-Yosef_etal_2018_Cognition}, to the modeling of detailed spatiotemporal interpretation, including modeling of internal parts and their spatiormeporal features and relations.
The spatiotemporal features and relations can be complex (e.g., a spatiotemporal relation of ‘parts containment’) to allow human-like detailed spatiotemporal interpretation.
Triggering these more complex features will therefore be done in a selective, top-down manner, and complementary to a first-stage forward aggregation of spatiotemporal filters (~\cite{Tran_etal_2015_ICCV,Hara_etal_2018_CVPR}). Focusing on a new type of models that cannot only label a particular action, but can also perform spatiotemporal interpretation will lead to a better understanding and more accurate modeling of spatiotemporal integration and human recognition. 
Full report is in ~\cite{Ben-Yosef_etal_2020_Cognition}. Code and data are available at 
\url{https://github.com/guybenyosef/introducing_minimal_videos.git}



\subsubsection*{Acknowledgments}
GBY, GK, and SU were supported by the National Science Foundation, 	Science and Technology Center Award CCF123121.  
SU was supported by EU Horizon 2020 Framework 785907 and ISF grant 320/16. 

\bibliography{gby_bib}
\bibliographystyle{iclr2020_conference}

\appendix
\section{supplementary material}
\begin{figure}
	\includegraphics[width=\linewidth]{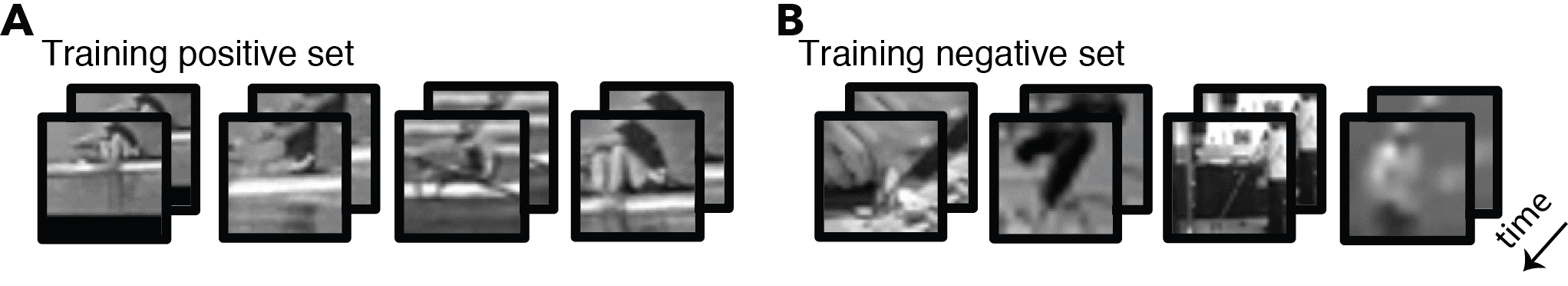}
	\caption{[supplementary]
		{\bf Positive and negative examples for training and testing classification models}. Positive examples are minimal videos from a single specific category and specific viewing position, while negative examples are non-category videos at the same style and resolution as minimal videos.}
	\label{fig:3D_2D_cnns}
\end{figure}


\end{document}